\documentclass[graybox]{svmult}

\usepackage[utf8]{inputenc}
\usepackage[T1]{fontenc}
\usepackage{mathptmx}        
\usepackage{helvet}          
\usepackage{courier}         
\usepackage{type1cm}
\usepackage{makeidx}
\usepackage{graphicx}
\usepackage{multicol}
\usepackage[bottom]{footmisc}
\usepackage{amssymb,amsmath}
\usepackage{bm}              
\newcommand{\llb}{[\![}
\newcommand{\rrb}{]\!]}
\usepackage{listings}
\usepackage{xcolor}
\usepackage{booktabs}
\usepackage{tikz}
\usetikzlibrary{arrows.meta,positioning,fit,backgrounds,calc}
\usepackage{url}
\usepackage{xurl}            

\usepackage{marginnote}
\usepackage{ragged2e}
\newif\ifshowcomments \showcommentsfalse
\newif\ifshowoldcb \showoldcbfalse
\newcommand{\CBold}[1]{\ifshowoldcb\CBold{#1}\fi}

\usepackage[backend=biber,style=numeric,sorting=nyt,giveninits=true,%
            maxbibnames=5,doi=false,url=false,isbn=false,eprint=false]{biblatex}
\AtEveryBibitem{\clearfield{note}\clearfield{issn}\clearfield{pagetotal}\clearfield{month}}
\newcommand{\obox}{\bm{\square}}
\newcommand{\oimp}{\mathbin{\bm{\rightarrow}}}
\newcommand{\ofa}{\bm{\forall}}
\newcommand{\oex}{\bm{\exists}}
\newcommand{\oor}{\mathbin{\bm{\vee}}}
\newcommand{\val}[1]{\lfloor #1\rfloor}

\lstdefinestyle{isa}{
  basicstyle=\ttfamily\scriptsize,
  mathescape=true,
  breaklines=false,
  keywordstyle=\bfseries,
  commentstyle=\rmfamily\itshape\color{black!55},
  morecomment=[s]{(*}{*)},
  keywords={definition,lemma,theorem,typedecl,type_synonym,consts,where,
            unfolding,by,using,assumes,shows,nitpick,sledgehammer,oops,
            axiomatization,abbreviation,fixes,proof,qed,then,have,show,and,
            datatype,fun},
  literate={==>}{{$\Longrightarrow$}}2 {=>}{{$\Rightarrow$}}2
           {-->}{{$\longrightarrow$}}2 {ALL}{{$\forall$}}1 {EX}{{$\exists$}}1
           {\%}{{$\lambda$}}1 {[|}{{$\llb$}}1 {|]}{{$\rrb$}}1
           {~}{{$\neg$}}1 {\&}{{$\wedge$}}1 {sigma}{{$\sigma$}}1 {tau}{{$\tau$}}1,
  columns=fullflexible,
  keepspaces=true,
  showstringspaces=false,
  frame=tb,
  framerule=0.4pt,
  framesep=6pt,
  xleftmargin=0.5em,
  aboveskip=0.9em,belowskip=0.4em,
}

\usepackage[hidelinks]{hyperref}   
\makeindex

\begin{document}

\title*{Mathematical Proof Assistants for \texorpdfstring{\\}{ }Teaching Logic: The LogiKEy Methodology}
\titlerunning{Proof Assistants for Teaching Logic: LogiKEy}
\author{Christoph Benzm\"uller, David Fuenmayor, and Luca Pasetto}
\authorrunning{C. Benzm\"uller, D. Fuenmayor, L. Pasetto}
\institute{%
Christoph Benzm\"uller \at Otto-Friedrich-Universit\"at Bamberg, Germany, and
Freie Universit\"at Berlin, Germany, \email{christoph.benzmueller@uni-bamberg.de}
\and David Fuenmayor \at Otto-Friedrich-Universit\"at Bamberg, Germany,
\email{david.fuenmayor@uni-bamberg.de}
\and Luca Pasetto \at University of Luxembourg, Luxembourg,
\email{luca.pasetto@uni.lu}}

\maketitle

\abstract*{We report on an approach to teaching logic to mixed groups of computer science,
mathematics, and philosophy students, based on the logico-pluralistic
\emph{LogiKEy} methodology, used for more than a decade in courses, summer
schools, and tutorials. LogiKEy uses classical higher-order logic (HOL) as a
universal \emph{meta}-logic in which object logics, classical and non-classical
alike, are encoded by defining their semantics; through these \emph{semantical embeddings} a single
proof assistant (e.g.~Isabelle/HOL), with its automated theorem provers and
(counter\mbox{-})model finders, becomes one environment in which students learn,
experiment with, and compare logics. After
making the pedagogical case for proof assistants in the logic classroom, we
present a graded sequence of classroom examples, each transition motivated by
a limitation of the preceding representation, by a need for more explicit
modelling resources, or by a new application. A liars-and-truth-tellers puzzle
leads from propositional to modal logic; the Wise Men puzzle leads on to
dynamic epistemic logic; Boolos's curious inference illustrates what a
higher-order meta-logic buys, even for automated proof search; Chisholm's
paradox takes the sequence into deontic logic, and from standard to dyadic
deontic logic; and G\"odel's ontological argument brings it to a research-level
metaphysical argument. We then rebut the objection that embedding everything in
classical HOL is monism rather than pluralism, reflect on three years of
teaching such a course, and sketch the portability of the approach beyond
Isabelle.}

\abstract{We report on an approach to teaching logic to mixed groups of computer science,
mathematics, and philosophy students, based on the logico-pluralistic
\emph{LogiKEy} methodology, used for more than a decade in courses, summer
schools, and tutorials. LogiKEy uses classical higher-order logic (HOL) as a
universal \emph{meta}-logic in which object logics, classical and non-classical
alike, are encoded by defining their semantics; through these \emph{semantical embeddings} a single
proof assistant (e.g.~Isabelle/HOL), with its automated theorem provers and
(counter\mbox{-})model finders, becomes one environment in which students learn,
experiment with, and compare logics. After
making the pedagogical case for proof assistants in the logic classroom, we
present a graded sequence of classroom examples, each transition motivated by
a limitation of the preceding representation, by a need for more explicit
modelling resources, or by a new application. A liars-and-truth-tellers puzzle
leads from propositional to modal logic; the Wise Men puzzle leads on to
dynamic epistemic logic; Boolos's curious inference illustrates what a
higher-order meta-logic buys, even for automated proof search; Chisholm's
paradox takes the sequence into deontic logic, and from standard to dyadic
deontic logic; and G\"odel's ontological argument brings it to a research-level
metaphysical argument. We then rebut the objection that embedding everything in
classical HOL is monism rather than pluralism, reflect on three years of
teaching such a course, and sketch the portability of the approach beyond
Isabelle.}
\section{The Logic Pluralism Challenge in Logic Education}
\label{sec:intro}

Logic is a mandatory, and notoriously demanding, subject in the curricula of
several disciplines at once. In computer science it underpins verification,
knowledge representation, and the theory of computation; in mathematics it
supplies the language of proof; in linguistics it underpins the formal
semantics of natural language; and in philosophy it is at once a tool and an
object of study.
Yet these audiences arrive with very different backgrounds,
motivations, and tolerances for formalism. Teaching logic well is therefore,
first of all, an \emph{interdisciplinary} challenge: the same room may hold a
student who programs fluently but has never written a proof, one who writes
rigorous proofs but has never seen them checked by a machine, and one who
argues subtly in natural language but freezes at a quantifier~\cite{C58,C90}.

A second challenge concerns \emph{content}. The applications that make logic
attractive today (artificial intelligence, natural-language semantics, the
formal analysis of law and ethics) routinely require reasoning \emph{beyond}
classical first-order logic: modal, deontic, epistemic, temporal, and many
further non-classical logics, often in combination. The traditional response
is to teach each such logic with its own syntax, its own proof calculus, and,
where tool support exists at all, its own software. For a mixed and
time-limited student audience this fragmentation is costly: every new
logic looks like a fresh start, and the conceptual thread (what all these
logics have in common, and how they differ) is easily lost.

This chapter reports on an approach that addresses both challenges together.
Its guiding idea is to fix a single, sufficiently expressive
\emph{meta-logic}, classical higher-order logic (HOL), and to obtain the
many object logics of interest \emph{inside} it, by means of
\emph{semantical embeddings}: encodings that define, within HOL, the semantics
of each object logic.\footnote{``Semantical embedding'' is the term established
in the LogiKEy literature; what is embedded into HOL is the \emph{semantics} of
the object logic. Despite the surface similarity, the term is unrelated to the
vector ``embeddings'' of machine learning. Sect.~\ref{subsec:sse} explains the
technique by example.}
A single proof assistant, in our courses Isabelle/HOL, then
provides one uniform working environment, complete with automated theorem
provers and (counter\mbox{-})model finders, in which a heterogeneous student group
can learn, experiment with, and, crucially, \emph{compare} logics. This is
the pedagogical face of the \emph{LogiKEy} methodology for logic and
knowledge engineering~\cite{J48,J65,J40,C90}.

Our thesis is that such a platform lets a mixed cohort learn \emph{many}
logics uniformly, with immediate machine-checked feedback, along a single
path that runs from elementary logic puzzles up to research-grade arguments
in ethics and metaphysics. In the vocabulary of this volume, the approach
answers its three organising questions at once. \emph{How} to teach logic: by
machine-supported, exploratory practice. \emph{What tools} to use: a
general-purpose proof assistant repurposed for teaching. \emph{What} to teach:
a plurality of logics, presented uniformly rather than in isolation. The last
point connects our proposal to the broader case, made elsewhere in this
volume, for teaching logic in a pluralistic and application-aware
spirit;\footnote{See, in the ``Philosophy'' part of this volume, the
contributions on teaching logic as formal philosophy and on teaching
non-classical logics through the Liar paradox, with which our computational
proposal is in close sympathy.} what we add is a concrete, executable way to
realise it in the classroom.

In sum, this chapter contributes (i)~a pedagogical case for teaching many
logics uniformly within a proof assistant; (ii)~a graded sequence of classroom
examples, from a liars-and-truth-tellers puzzle to G\"odel's ontological
argument, in which each transition is motivated by a limitation exposed in
the preceding representation, by a need for more explicit modelling resources,
or by a new application, so that students examine different reasons for
changing a logic or its formalisation.
It further contributes (iii)~an account of the
supporting tools, including automated assessment and a Kripke-model
visualiser, and (iv)~a reflection on three years of teaching such a course to
a mixed audience.

A word on what is, and is not, new. LogiKEy itself is an established research
programme, and a two-page abstract under the same title~\cite{C90} already
sketched the idea of teaching with it. What is new here is the worked course and its rationale (Sect.~\ref{sec:ladder}), the tools
that make it usable by non-specialists (Sect.~\ref{sec:tools}), the positioning
of the meta-logical stance against the charge of monism
(Sect.~\ref{subsec:pluralism}), and a sketch of portability to other proof
assistants (Sect.~\ref{sec:outlook}). Underlying all of these is one thread we
ask the reader to follow: what is gained is not simply that a proof assistant
is used, but that a single \emph{expressive meta-logic} is. Its expressiveness
buys an economy of formalisation, since each further object logic is a set of
definitions rather than a new system; that economy in turn buys automation,
since every embedded logic inherits the same provers and model finders; and
only on top of these does the teaching application rest. Boolos's inference
(Sect.~\ref{subsec:boolos}) shows that the first of these steps can pay off
even where one would least expect it. The chapter belongs to the
\emph{Programmes} part of this volume: it describes a methodological
infrastructure and a concrete teaching application, not an empirical study of
learning.

We write for logic educators and for readers from philosophy, mathematics, and
computer science who have not necessarily used a proof assistant. A basic
acquaintance with propositional and modal logic suffices. The Isabelle/HOL
listings are illustrative and can be read as annotated pseudo-code; no
knowledge of Isabelle, type theory, or programming is assumed, and the same
holds for the students in the course described below. Adopting the
infrastructure is a different matter: instructors will find the complete,
machine-checked sources openly available (Sect.~\ref{sec:outlook}), but putting
them to work does require an initial investment in the tools.

\section{The LogiKEy Idea in One Picture}
\label{sec:idea}

The whole approach rests on one central idea, which we now explain once so
that the later sections can take it for granted. The idea has three
ingredients: a
meta-logic, an embedding technique, and a layered architecture. We keep the
treatment deliberately light, introducing only what the classroom examples of
Sect.~\ref{sec:ladder} require: enough to follow the proofs and countermodels
that follow.

\subsection{Higher-order logic as a universal meta-logic}
\label{subsec:hol}

Classical higher-order logic, Church's simple type theory~\cite{B27}, is
ordinary classical logic with two decisive additions. One may quantify not only
over individuals but also over functions and predicates, which makes HOL
expressive enough to \emph{talk about} other logics: their formulas, their
notions of truth, their models. And one may form functions on the fly by
$\lambda$-abstraction, which makes encodings compositional and elegant and
gives them the flavour of elementary functional programming: a modal
connective, for instance, simply becomes a function on world-dependent
propositions (Sect.~\ref{subsec:sse}). LogiKEy exploits both, using HOL as a
\emph{universal meta-logic}~\cite{J40}. Students need no prior
acquaintance with type theory to use it; in practice they meet HOL simply as
the language of the proof assistant, and its higher-order features appear
exactly when, and only when, a particular object logic calls for them.

\subsection{Semantical embeddings, by example}
\label{subsec:sse}

The technique by which LogiKEy turns HOL into a host for other logics is
\emph{shallow semantical embedding} (SSE)~\cite{J40,%
C98}; a deep alternative is discussed in Sect.~\ref{subsec:pluralism}. The idea is best seen on propositional modal
logic. Read a modal formula not as an opaque symbol but as a
\emph{predicate on possible worlds}: ``$\varphi$ holds'' becomes
``$\varphi$ holds \emph{at a world} $w$''. Necessity is then nothing but
truth at all accessible worlds. Writing $\sigma$ for the type of such
world-predicates (the listings use the same name) and $R$ for the
accessibility relation, the modal connectives are \emph{defined} in HOL by
\begin{align}
  \bm{\neg}\varphi      &\;:=\; \lambda w.\; \neg\,(\varphi\, w), &
  \varphi\oimp\psi   &\;:=\; \lambda w.\; \varphi\, w \longrightarrow \psi\, w, \notag\\
  \obox\varphi      &\;:=\; \lambda w.\; \forall v.\; R\,w\,v \longrightarrow \varphi\, v, &
  \val{\varphi} &\;:=\; \forall w.\; \varphi\, w, \label{eq:sse}
\end{align}
where $\val{\varphi}$ (``$\varphi$ is valid'') abbreviates truth at every
world. Wherever the two levels meet in one formula, as they do here, we set the
\emph{object-level} connectives in bold: the $\bm{\neg}$ and $\obox$ on the
left are being defined, the $\neg$, $\longrightarrow$ and $\forall$ on the
right are HOL's own. The notation $\lambda w.\,\ldots$ denotes the function that
assigns to each world $w$ the truth value described after the dot; so
$\neg\varphi$, for instance, is the proposition that is true at exactly those
worlds at which $\varphi$ is false. That is the \emph{entire} definition of the modal logic
$\mathbf{K}$: everything else (the axiom $K$, the necessitation rule, and
all theorems) is a \emph{consequence} of these definitions, obtained by
ordinary reasoning in HOL rather than stipulated. In Isabelle/HOL the definitions
of~\eqref{eq:sse} are a handful of lines, after which one can already state
and prove modal facts (Fig.~\ref{fig:sse-snippet}).\footnote{Throughout, the
Isabelle/HOL code is shown as Isabelle displays it, with the usual logical
symbols, and occasionally deliberately simplified for a non-specialist
reader. The faithful,
machine-checked sources are openly available (cf.\ Sect.~\ref{sec:outlook}).}
Because the embedding is
so direct, an off-the-shelf HOL prover reasons \emph{in the modal logic} for
free; just as importantly, a HOL (counter\mbox{-})model finder produces modal
\emph{countermodels} when a conjecture fails. (What such a countermodel looks
like, first as the model finder prints it and then as a picture, is shown in
Figs.~\ref{fig:nitpickraw} and~\ref{fig:kripke} below.)

\begin{figure}[t]
\begin{lstlisting}[style=isa]
(* axiom K holds in every frame (a consequence of the definitions): *)
lemma K:  "$\val{\obox(p \oimp q) \oimp (\obox p \oimp \obox q)}$"  by simp
(* correspondence: the 'T' schema holds exactly on reflexive frames: *)
lemma Tc: "(ALL p. $\val{\obox p \oimp p}$) = (ALL x. R x x)"  by metis
(* without reflexivity, Nitpick refutes the schema: *)
lemma     "$\val{\obox p \oimp p}$"  nitpick
\end{lstlisting}
\caption{Working with the embedding~\eqref{eq:sse}. Axiom $K$ is proved in
every frame. The schema $\obox\varphi\oimp\varphi$ is shown to correspond
\emph{exactly} to reflexivity of the accessibility relation, and without
reflexivity Nitpick returns a countermodel: proof and refutation are the same
one-line gesture. (The plain $\forall$ and \texttt{R} are meta-level; the
bold $\obox$ and $\oimp$ are the \emph{embedded} object connectives. \texttt{simp} and \texttt{metis} are automatic proof methods; they
do construct proofs, which Isabelle checks but does not normally display, and
any of them can be replaced by a step-by-step proof in Isabelle's Isar
language.)}
\label{fig:sse-snippet}
\end{figure}

The pedagogical dividend appears the moment one varies the frame. Imposing a
condition on $R$ selects a stronger logic: reflexivity yields $\mathbf{KT}$,
seriality yields $\mathbf{KD}$ (the base of \emph{deontic} logic, of which
more in Sect.~\ref{sec:ladder}), transitivity yields $\mathbf{K4}$, and so
on up to $\mathbf{S5}$. Students do not take these correspondences on
faith: they \emph{test} them, and watch the same machinery deliver a
different logic each time. One meta-logic, many object logics: this refrain
returns throughout the chapter.

\subsection{A layered architecture---and why it helps}
\label{subsec:layers}

\begin{figure}[t]
\centering
\begin{tikzpicture}[font=\scriptsize,
  layer/.style={draw,rounded corners,minimum width=6.4cm,
                minimum height=0.5cm,align=center},node distance=1.6mm]
  \node[layer,fill=green!12]  (l3) {L3 --- Applications};
  \node[layer,fill=orange!14,below=of l3] (l2) {L2 --- Domain-Specific Languages, Theories, Ontologies};
  \node[layer,fill=yellow!18,below=of l2] (l1) {L1 --- Object Logic(s) and their Combinations};
  \node[layer,fill=blue!12,below=of l1]  (l0) {L0 --- Meta-Logic (HOL)};
  \begin{scope}[on background layer]
    \node[draw,rounded corners,fit=(l3)(l0),inner sep=2mm,fill=gray!6] (box) {};
  \end{scope}
  \node[rotate=90,anchor=south,font=\scriptsize\bfseries] at ([xshift=-1.5mm]box.west) {LogiKEy};
\end{tikzpicture}
\caption{The four layers of the LogiKEy methodology. Object logics are
embedded in HOL (L0$\to$L1), used to build domain theories (L2), and applied
to concrete scenarios (L3). Students may enter at the layer matching their
discipline.}
\label{fig:layers}
\end{figure}
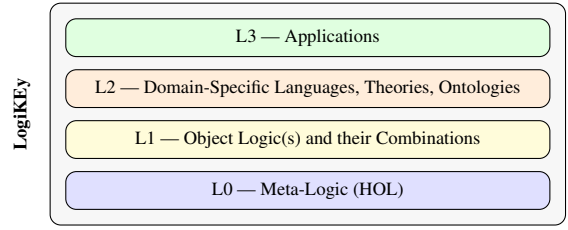

LogiKEy organises this idea into four layers (Fig.~\ref{fig:layers}).
At the base (L0) sits the meta-logic, HOL. On top of it one encodes object
logics and their combinations (L1). Using those logics one formulates
domain-specific languages, theories, and ontologies (L2). With these in hand
one models and assesses concrete application scenarios (L3)~\cite{J48,W72}.\footnote{In the framework paper~\cite{J48} the three
\emph{engineered} layers L1--L3 rest on the meta-logic HOL as a substrate; we
simply number that substrate L0.} For teaching,
the layering has a pleasant consequence. Students can enter at the layer that
suits their background, a philosopher at L2/L3 with an ethical scenario, a
computer scientist at L0/L1 with the embedding machinery, and still meet on
the same platform and speak the same underlying language. The architecture
thus doubles as a map of an interdisciplinary division of labour.

The tools the student actually touches are few. \emph{Isabelle/HOL}~\cite{NipkowPaulsonWenzel2002} is the surface: a mature, well-documented
proof assistant with a modern editor. Behind a single command,
\emph{Sledgehammer}~\cite{BlanchetteEtAl2016} invokes external automated
theorem provers to \emph{find} proofs, while \emph{Nitpick}~\cite{BlanchetteNipkow2010} searches for (counter\mbox{-})models. For the learner
these are a proof-finding and a refutation companion
(Sect.~\ref{sec:tools}). For interactive proof, Isabelle offers, among
various other proof languages, \emph{Isar}: a declarative language modelled on
the way mathematicians argue, in which a machine-checked proof reads like a
structured argument.

\subsection{Semantical embeddings and logical pluralism}
\label{subsec:pluralism}

The embedding technique is not new. Shallow semantical embeddings grew out of
the automation of multimodal and quantified multimodal logics in simple type
theory~\cite{B7,J26}; for an overview see~\cite{J40}. A
companion \emph{deep} style, in which the object syntax is represented as a
datatype with an explicit interpretation function, has been worked out
alongside it. For simple logics such as propositional modal logic it has
been proved by machine that the two styles are \emph{faithful}: they prove
exactly the same formulas, so the object logic is captured
correctly~\cite{C98}. (That paper was itself written with logic education in
mind.) For teaching, the shallow style is the natural default: it is short,
readable, and lets the prover work directly on object-level formulas
(Fig.~\ref{fig:shallowdeep} shows the difference on the box). A further
dividend of embedding a logic, rather than hard-coding it, is that one reasons
at two levels at once. The same machinery that proves object-level theorems
also establishes facts \emph{about} the logic, such as the frame
correspondences of Sect.~\ref{subsec:sse}, and finds countermodels when either
fails.

\begin{figure}[t]
\begin{lstlisting}[style=isa]
(* shallow: the box is a HOL definition on world-predicates (type sigma = i => bool) *)
definition box :: "sigma => sigma" ("$\obox$")  where "$\obox$ p = (%w. ALL v. R w v --> p v)"
(* deep: formulas are a datatype, evaluated by an explicit interpretation function *)
datatype form = Atom nat | Neg form | Imp form form | Box form
consts val :: "nat => i => bool"      (* valuation of the atoms, a constant like R *)
fun eval :: "form => i => bool" where
  "eval (Atom n) w = val n w"
| "eval (Box f) w = (ALL v. R w v --> eval f v)" | ...
\end{lstlisting}
\caption{The two styles of embedding, on the box (schematic; the deep
evaluation function is shown for the atom and box cases only). In the \emph{shallow} style
the object connective simply \emph{is} a HOL function on world-predicates, so
the prover works on object formulas directly. In the \emph{deep} style object
formulas are data, and a separate evaluation function assigns them a meaning;
this makes the syntax itself an object of study, at the price of one more
level of indirection. Faithfulness means that both styles validate exactly the
same formulas~\cite{C98}.}
\label{fig:shallowdeep}
\end{figure}

Conceptually, the method turns \emph{logical pluralism} from a slogan into a
practice. Instead of treating each logic as a self-contained system with its
own calculus, one exhibits many logics as definitional variants within a single
framework, and can compare them, combine them, and move between them without
changing tools. This computational pluralism sits naturally beside the
philosophical case, made elsewhere in this volume, for teaching logic as formal
philosophy and for taking a plurality of logics seriously in the curriculum; it
offers a concrete, executable counterpart to those arguments.

A philosopher will rightly press the obvious objection: if every object logic is
interpreted in \emph{classical} HOL, is this not in truth a classical
\emph{monism} in disguise, and the ``pluralism'' merely cosmetic? The reply we
and others give~\cite{B30} distinguishes a \emph{methodology} from a
\emph{foundation}. LogiKEy does not enthrone HOL as the one true logic in which
all reasoning must be conducted: the choice of HOL rests on pragmatic
grounds (its expressiveness and its mature automation) and is not constitutive
of the method. What \emph{is} constitutive is that the object logic, and to some
extent even the meta-logic, are kept negotiable and made objects of study. A
single rich foundation, by contrast, silently embeds commitments, such as
excluded middle, the axiom of choice, existential import, or extensionality,
that remain
opaque to the user and against which any dissenter must work: \emph{a foundation
forecloses, whereas a methodology explores}~\cite{B30}. This exploration pays
off in practice: when Zalta's object theory was encoded as an object logic in
HOL, the interaction with Isabelle/HOL brought to light a paradox that had been
inadvertently reintroduced and had previously gone undetected~\cite{J49,B30}. A
smaller example, taken from Rung~4 below, shows how such a commitment enters a
formalisation. Do the quantifiers of the object logic range over all possible
individuals, or only over those that actually exist at a world? This choice
between a \emph{possibilist} and an \emph{actualist} reading is invisible in
the informal text of G\"odel's ontological argument. In the encoding it is a single
definitional decision, and which results remain provable, and under which
frame conditions, depends on it~\cite{C65,J70}.
Making the choice explicit is what allows it to be varied and discussed in
class. The stance is thus pluralism set deliberately against a \emph{logical
imperialism} (the term is from~\cite{B30}) that would impose one logic on every
problem. In the classroom it has a welcome corollary: no student's
``home'' logic is crowned the winner, and the differences \emph{between}
logics, not one orthodoxy, become the subject matter.

\section{Why a Proof Assistant in the Logic Classroom}
\label{sec:why}

Before turning to concrete classroom material, we set out the pedagogical
case for using an interactive proof assistant to teach logic. Our experience
builds on an earlier line of work on theorem provers in teaching~\cite{C58},
and our courses follow the Humboldtian ideal of a unity of research and
teaching: they aim at a \emph{research-oriented} education that can culminate
in students reproducing current research results~\cite{C58}.

A word on the status of the claims that follow. Three kinds of claim run
through this chapter. \emph{Formal results} (that a given formula is provable
or refutable) are machine-checked and reported as such in
Sect.~\ref{sec:ladder}. \emph{Methodological} claims (that one framework can
carry all of these logics uniformly) are borne out by the worked examples. The
\emph{pedagogical} claims of this section, and the classroom observations of
Sect.~\ref{sec:experience}, rest on our own teaching experience: three runs of
one course with small, self-selected cohorts, and earlier courses in the same
line. We offer them as considered impressions and as hypotheses worth testing,
not as findings of a controlled study.

\runinhead{Immediate, machine-checked feedback---no hidden gaps.}
On paper, the reasoning steps a student finds ``obvious'' are often the
ones that hide errors, and feedback arrives days later, if at all. In a proof
assistant every step is checked as it is made; a gap cannot be waved away,
because the system will not accept it. The effect, in our experience, is not
punitive but clarifying: students learn where an inference's real content lies.

\runinhead{Logic as an experimental subject.}
With proof search and model finding one command away, students can
\emph{try things}: Does this formula hold on this class of frames? What
happens if I drop seriality? Which axioms separate modal logic $\mathbf{S4}$ from
$\mathbf{S5}$? The proof assistant turns questions that would otherwise be
rhetorical into experiments with immediate answers, encouraging a habit of
conjecture-and-test that is closer to mathematical practice than to
rote calculation~\cite{C90}.

\runinhead{What students actually do.}
Sessions begin with small, complete, \emph{runnable} theories that students
execute and modify rather than read passively: change an axiom and watch a
theorem turn into a countermodel; strengthen a frame condition and watch the
countermodel turn back into a proof. The core task is \emph{formalisation}:
rendering an informal statement (a puzzle, a philosophical principle, a legal
rule) in a chosen object logic and asking the machine whether the intended
conclusions follow. This targets the skill a mixed audience most needs and most
lacks, namely moving carefully between natural language and formal notation,
now with an oracle to catch mistakes (and, more recently, with language
models to draft a first attempt; see Sect.~\ref{sec:tools}). Assignments are
correspondingly
open-ended (``model this scenario and decide whether\ldots'', ``find a
countermodel showing that\ldots'') rather than narrow calculations, because
automatic checking makes such tasks tractable both to set and to assess. This
iterative loop of formalising, testing, and revising has a philosophical side
as well: it is a form of \emph{computational hermeneutics}, in which the
meaning of an informal concept is made explicit precisely by the succession of
formalisations the machine accepts and rejects~\cite{B18}.

\runinhead{A dialogical style of learning.}
Used this way, the system becomes a partner in a dialogue between the learner
and the material: pose a claim, attempt a proof, inspect a countermodel,
revise. This fits the model of \emph{dialogical learning} of Ruf and Gallin,
in which a core idea, an open-ended commission, a learning journal, and
feedback chain production and reception into an iterative loop~\cite{C58}. A
proof assistant is unusually well suited to play the machine side of this
dialogue. An Isabelle theory accumulated over a term is a \emph{formally
verified} learning journal: a record of a student's reasoning in which every
accepted step is guaranteed correct, and in which even unsuccessful attempts
can be preserved and annotated for later reflection. Exemplary passages from
such journals can be curated into a shared collection and fed back to the
class as new learning material~\cite{C58}, and since Isabelle typesets a theory
into a PDF, a journal doubles as a publishable, machine-checked document. The
division of labour in this dialogue is worth stating plainly: the assistant
supplies reliable formal feedback, and interpreting that feedback, connecting
it to the logical or philosophical question behind it, remains the student's
work.

\runinhead{Lowering the barrier---and the affective payoff.}
A recurring worry about formal methods in teaching is that the tooling itself
becomes the obstacle. Two features mitigate this. First, one starts from very
small, self-contained examples and grows them; nothing must be built from
scratch. Second, the machine's willingness to \emph{refute} a plausible
conjecture, and to reward a second or third attempt at a proof, seems to us to
encourage the dispositions that a virtue-oriented view of logic education
prizes: open-mindedness, humility, persistence. It may also help defuse the
``logic anxiety'' that formal notation provokes in humanities-trained
students. Both are impressions from our courses rather than measured
effects.\footnote{Both themes are taken up elsewhere in
this volume, in the contributions on the epistemology of virtues and on
logic anxiety, respectively.}

\section{A Course Walkthrough: The Graded Ladder}
\label{sec:ladder}

We now show the methodology in action, as a graded progression of classroom
examples that students climb like the rungs of a ladder: each rung is a new
logic, reached from the one below it.
The concrete setting is a lecture course with tutorial, \emph{Universal
Logic and Universal Reasoning} (\texttt{AISE-UL}), taught by two of us at the
University of Bamberg in three successive winter terms (2022/23, 2023/24,
2024/25). It is a $2{+}2$~SWS, $6$~ECTS course for computer-science
students, offered to philosophy students as a $3$~ECTS seminar
variant~\cite{C90}. It continues a line of teaching that runs
through the first author's \emph{Universal Logical Reasoning} course at the
Freie Universit\"at Berlin (from winter 2018/19) back to the
\emph{Computational Metaphysics} course, which won the university's central
teaching prize for 2015 and ran in the summer of 2016~\cite{R57}.\footnote{The
prize, awarded to teaching concepts to be realised in the following academic
year, was announced on 1 February 2016; see the FU Berlin press release
No.~022/2016, \url{https://www.fu-berlin.de/presse/informationen/fup/2016/fup_16_022-lehrpreis-benzmueller-logik-gottesbeweis/index.html}.}
That course was designed and taught together with Max Wisniewski and Alexander
Steen~\cite{R57,C58}; Steen went on
to become a contributor to and developer of LogiKEy, and he now pursues the
same line, with a stronger emphasis on proof automation, as junior professor
at the University of Greifswald. The course has also yielded what this chapter
hopes for: two of its participants went on to write their theses, up to the
doctoral level, in the same line. One began with a master's thesis
mechanising Zalta's \emph{Principia Logico-Metaphysica}, the work in which the
paradox recounted in Sect.~\ref{subsec:pluralism} came to
light~\cite{KirchnerMSc2017,J49}, and completed a doctorate on
computer-verified foundations of metaphysics~\cite{KirchnerPhD2022}. The other
is the second author of this chapter. He began with a bachelor's thesis that
applied the method to Lowe's modal ontological
argument~\cite{FuenmayorBSc2018,J37}. A series of publications followed, on
the ontological argument, on computational hermeneutics, and on the normative
applications and the Boolos example on which this chapter
draws~\cite{C65,B18,C80,J65,J61}, before a doctorate on a universal formal
language for argumentative reasoning agents~\cite{FuenmayorPhD2025}. The name \emph{LogiKEy}, and the
methodology's international spread, date from the first author's years at the
University of Luxembourg, where Xavier Parent and Leendert van der Torre became
convinced of its virtues and joined its development~\cite{J48}. The
methodology reaches well beyond this one course: it has been used in graduate
courses on computational metaphysics and on intelligent agents (the latter at
the University of Luxembourg), in invited courses and an ongoing research
collaboration at Zhejiang University in China,\footnote{The invited lecture
course \emph{Universal Logical Reasoning} (2020, part of a course on Logic in
AI; \url{https://www.xixilogic.org/events/2020/08/course-logic-in-ai/}) and
the foreign-expert projects at ZLAIRE
(\url{https://www.zlaire.net/foreignexperts/}).} at summer schools (from a
first tutorial on higher-order modal logics, given with Bruno Woltzenlogel
Paleo at the Reasoning Web Summer School and the ANU Logic Summer School in
2015, to the ESSLLI 2026 course \emph{Experimenting with the LogiKEy Framework
\& Methodology} in Prague\footnote{See
\url{https://logikey.org/CoursesAndTutorials/2026-ESSLLI/}.}), and in numerous
invited tutorials and lecture courses over the past decade, among them at the
Berkeley--Stanford Circle in Logic and Philosophy, at the CIMI thematic
trimester in Toulouse, and in Brazil.\footnote{See \url{https://christoph-benzmueller.de/teaching.html} for
the full record of LogiKEy-based courses and tutorials.} Enrolments in
such advanced courses are naturally small, and the audience is genuinely
mixed: in Bamberg a handful of computer-science and philosophy students, in
the earlier Berlin courses also mathematics students, working side by side on
the same platform.

A semester follows the ladder. The first weeks establish HOL and the proof
assistant and settle the propositional warm-up, alongside small exercises (such
as a compact HOL proof of Cantor's theorem~\cite{R74}) that build fluency with
the tool. The middle of the course introduces the embedding technique and
works through modal, then epistemic and deontic logics. The final weeks are
given to the capstone and to student projects. Prerequisites are deliberately light:
a basic acquaintance with propositional and a little modal logic, and \emph{no}
programming experience. This is what makes a genuinely mixed enrolment
possible.

Each transition is motivated by a limitation exposed in the preceding
representation, by a need for more explicit modelling resources, or by a new
application: the student moves to a new logic not because the syllabus
prescribes it, but because an experiment, or a new question, shows what the
old one cannot do. The sequence thus lets students examine different reasons
for changing a logic or its formalisation, and each example exposes a
different facet of the methodology. Table~\ref{tab:ladder}
lays the path out at a glance. It is the actual backbone of the Bamberg course,
but we present it as an \emph{exemplary} path rather than a prescribed
curriculum: the steps differ in kind, and an instructor may enter, leave, or
branch at any rung. Throughout, the refrain ``one meta-logic, many object
logics, always machine-checked'' becomes a matter of necessity rather than of
sightseeing.

\begin{table}[t]
\caption{The graded ladder at a glance: for each example, the object logic,
the problem that motivates it, the ingredient it adds, and the kind of step it
represents: a limitation exposed in the preceding representation, a need for
more explicit modelling resources, or a new application.}
\label{tab:ladder}
\small\RaggedRight\hbadness=10000
\setlength{\tabcolsep}{3pt}
\renewcommand{\arraystretch}{1.2}
\begin{tabular}{@{}p{1.75cm}p{1.9cm}p{2.6cm}p{2.6cm}p{2.0cm}@{}}
\toprule
Example & Object logic & Motivating problem & New ingredient & Kind of step \\
\midrule
Rung 1: Liars Street & propositional (HOL itself) & who lives where?; then: ``knows'' as a truth-function collapses & machine-checked reasoning; the layers L1--L3 & first contact \\
Rung 2: modal cube & modal $\mathbf{K}$ to $\mathbf{S5}$ & the extensional collapse & possible worlds, accessibility, frame conditions & limitation exposed \\
Sequel: Wise Men & multi-agent $\mathbf{S5}$ + PAL & announcements change what is known & announcement as model restriction & new modelling resources \\
Interlude: Boolos & HOL vs.\ FOL (meta-level) & what does higher-order buy? & comprehension as a cut & illustrates the meta-logic's virtues \\
Rung 3: Chisholm & SDL ($\mathbf{KD}$), then dyadic DL & contrary-to-duty obligations & deontic reading; countermodels as pictures; preference relations & new application; then limitation exposed \\
Rung 4: G\"odel & higher-order modal logic & necessary existence & quantification over properties; frame conditions per theorem & extension to a research argument \\
\bottomrule
\end{tabular}
\end{table}

\subsection{Rung 1---A logic puzzle, and its limits}
\label{subsec:rung1}

The course opens with a puzzle of the ``liars and truth-tellers'' kind. Two
children, Nilda and Carla, each live either in Liars Street (whose inhabitants
always lie) or on Truth-tellers' Road (whose inhabitants always tell the
truth); from what they say about one another, the student asks the machine which
worlds fit the story. The scenario is encoded directly in HOL, and the first aim is purely
experiential: to feel what it is like to reason with a partner that checks
every step. The partner answers either ``yes, and here is why'' (Sledgehammer
finds a proof) or ``no, and here is a counter-scenario'' (Nitpick exhibits a
model or a countermodel). The encoding (Fig.~\ref{fig:liars}) is instructive in a second
way, too: it is laid out in exactly the LogiKEy layers of
Sect.~\ref{subsec:layers}. The object logic is just HOL's own propositional
connectives (L1); the vocabulary ``\_~says~\_'' and ``\_~lives\_in~\_'',
together
with the two bridge axioms, forms a small domain-specific language (L2); and the
concrete question about Nilda and Carla is the application (L3). The puzzle is
thus also a first, painless glimpse of the architecture.

\begin{figure}[t]
\begin{lstlisting}[style=isa]
consts says :: "person => bool => bool" ("_ says _")   (* what an inhabitant asserts *)
axiomatization where
  Liars: "p lives_in Liars ==> (p says X) = (~X)"  (* liars invert the truth *)
  Truth: "p lives_in Road  ==> (p says X) =   X "  (* truth-tellers preserve it *)
(* Nilda and Carla each vouch that the other tells the truth. Which worlds fit? *)
lemma
  assumes "Nilda says (Carla lives_in Road)"
      and "Carla says (Nilda lives_in Road)"
  shows   "Nilda lives_in s1 & Carla lives_in s2"
  nitpick [satisfy]   (* Nitpick returns the models consistent with the story *)
\end{lstlisting}
\caption{The warm-up ``liars and truth-tellers'' puzzle, in outline (the declarations of
the types and of \texttt{lives\_in} are omitted). Rather than
work the knight-and-knave cases by hand, the student has
\texttt{nitpick[satisfy]} \emph{find} the configurations consistent with what
Nilda and Carla say: the ``answer models'', read off the free variables
\texttt{s1} and \texttt{s2}. Here two survive (both tell the truth, or both
lie); enumerating that solution space is exactly the machine's job.}
\label{fig:liars}
\end{figure}

So far, so propositional. The turn comes when we ask what ``\_ says \_'' and
``\_ knows \_'' actually \emph{mean}. In the first encoding they are ordinary
truth-functions: ``$X$ says $\varphi$'' is simply a truth value that depends
only on the truth value of $\varphi$. That suffices for the basic puzzle, but it conceals
a trap, and the course springs it deliberately (Fig.~\ref{fig:omniscience}, upper listing).
Because such an operator sees only the \emph{truth value} of its argument, any
two propositions that merely happen to be true become interchangeable inside it:
the moment Carla ``knows'' one truth, the machine proves that she ``knows''
every other truth as well. This \emph{extensional collapse} (the crudest member
of the family of \emph{hyperintensionality} problems, of which the well-known
\emph{logical-omniscience} problem is a subtler relative) is not a caveat to be
issued in a lecture but a theorem the students derive from their own encoding.
The lesson is inescapable: a truth-functional, propositional reading of
``says'', ``knows'', and ``believes'' is untenable, and something more
expressive is required.

\begin{figure}[t]
\begin{lstlisting}[style=isa]
consts knows :: "person => bool => bool" ("_ knows _")   (* a plain truth-function *)
(* P and Q are both true, and Carla knows P; then she 'knows' Q as well: *)
lemma "[| P; Q; Carla knows P |] ==> Carla knows Q"  by simp
\end{lstlisting}
\smallskip
\begin{lstlisting}[style=isa]
typedecl i                          (* the type of possible worlds *)
type_synonym sigma = "i => bool"    (* a proposition now depends on worlds *)
consts R :: "person => i => i => bool"    (* each agent's accessibility relation *)
definition Knows :: "person => sigma => sigma" ("_ knows _")
  where "a knows p = (%w. ALL v. R a w v --> p v)"   (* box, per agent *)
(* the same premises as above, now at a world w: the entailment fails *)
lemma "[| P w; Q w; (Carla knows P) w |] ==> (Carla knows Q) w"  nitpick
\end{lstlisting}
\caption{The extensional collapse, and its modal repair. \emph{Above}: because
``knows'' sees only the \emph{truth value} of its argument, and \texttt{P} and
\texttt{Q} are both true, Isabelle proves that knowing \texttt{P} forces
knowing \texttt{Q}; \emph{any} two truths become interchangeable, a reductio
the students derive for themselves. \emph{Below}: a proposition becomes
world-dependent, and ``knows'' becomes a box over the worlds the agent
considers possible (frame conditions, such as those of $\mathbf{S5}$, are
added afterwards). The collapse is gone: for the very same premises Nitpick
now finds a world \texttt{w} at which \texttt{P} and \texttt{Q} both hold and
\texttt{P} is known, yet \texttt{Q} is not, because \texttt{Q} fails at some
other world Carla considers possible. Logical omniscience \emph{proper},
knowledge closed under consequence, survives and motivates finer logics.}
\label{fig:omniscience}
\end{figure}

\subsection{Rung 2---Modal logic, and the modal-logic cube}
\label{subsec:rung2}

The repair is modal logic. A proposition ceases to be a bare truth value and
becomes a \emph{function from possible worlds to truth values}; its type
changes from \texttt{bool} to $\sigma$ (which abbreviates \texttt{i} $\Rightarrow$
\texttt{bool}, for \texttt{i} the type of \emph{possible worlds}). ``$X$ knows $\varphi$'' is
then read as ``$\varphi$ holds at every world $X$ considers possible'': the box operator
of Sect.~\ref{subsec:sse}, now indexed by the agent
(Fig.~\ref{fig:omniscience}, lower listing). Two propositions that happen to agree in truth
value at the current world need no longer be interchangeable inside a modality,
since they may differ at \emph{other} worlds; the extensional collapse is
undone, and Nitpick confirms that the very same entailment now fails. Here modal logic is
not introduced, it is \emph{needed}: the limitation that ended Rung~1 is what
motivates the move.

Honesty compels a caveat, and the caveat itself makes the pluralist point. The
modal reading removes the \emph{crude}, extensional collapse, but not the finer
\emph{logical-omniscience} problem: according to modal logic $\mathbf{S5}$ a knower still knows every
logical consequence of what they know, and every valid formula. Faithfully
modelling resource-bounded, non-omniscient agents calls for \emph{yet further}
logics (hyperintensional, awareness-based, or impossible-worlds accounts), each
of which the same meta-logical framework can host in turn. Every repair exposes
the next question, and none of them forces us to leave the platform: that, in a
sentence, is the pedagogical case for logical pluralism.

With one modality in hand, the natural next question is which principles it
obeys, and here the framework pays a further dividend. Using the embedding of
Sect.~\ref{subsec:sse}, students explore the ``modal-logic cube''~\cite{C47}: they impose conditions
on the accessibility relation one at a time and watch the logic change under
their hands.
Reflexivity validates $\Box\varphi\to\varphi$ (axiom $\mathit{T}$: what is
known is true). Seriality validates $\Box\varphi\to\Diamond\varphi$ (axiom
$\mathit{D}$), and transitivity $\Box\varphi\to\Box\Box\varphi$ (axiom
$\mathit{4}$: positive introspection). Reflexivity together with the
Euclidean condition (any two worlds accessible from a given world are
accessible from each other) yields $\mathbf{S5}$, a familiar logic of
idealised knowledge.\footnote{The systematic verification~\cite{C47} labels the
$\mathit{T}$- and $\mathit{4}$-axioms $M$ and $IV$; we use the conventional
names.} Each
correspondence is \emph{proved} when it holds and \emph{refuted by an explicit
countermodel} when it does not---the same button, opposite verdicts. (The
systematic development behind the cube machine-verifies all 15 of its logics
and the 25 inclusions among them~\cite{C47}.) The refrain is now concrete: the object logic is selected by adding a single line,
and the meta-logic never changes. The frame condition behind axiom $\mathit{D}$ is
\emph{seriality}, which is characteristic of the deontic logic taken up in
Rung~3.

Even this, the course is careful to add, is not the last word. Reading
``says'' as a static box still mishandles the \emph{dynamics} of
announcement (what changes when something is publicly said), and nested
announcements soon outrun plain modal $\mathbf{K}$. That is the doorway to
\emph{dynamic epistemic logic}, which we take up in the worked example that
follows.

\subsection{An epistemic sequel: the Wise Men}
\label{subsec:wisemen}

The gap left open by Rung~1 was the \emph{dynamics} of communication, and the
classic vehicle for it is the Wise Men puzzle (equivalently, the ``muddy
children''). A king marks each of three wise men with a spot on the forehead,
white or black---in fact all three white; each sees the others' spots but not
his own. The king announces publicly that at least one spot is white, and then
asks, in turn, whether each wise man knows the colour of his own. The first two answer ``no'';
on hearing them, the third answers ``yes''---correctly. The puzzle is a staple
precisely because the decisive information is carried not by what is said about
the spots but by the \emph{announcements themselves}: each ``I do not know'' is a
public event that changes what everyone knows.

Static modal logic has no direct way to capture this: a fixed accessibility
relation cannot represent the \emph{update} an announcement performs, so each
announcement would have to be handled by re-encoding the model by hand.
\emph{Public-announcement logic} (PAL) supplies an operator for exactly
this.\footnote{In sheer expressive power PAL without common knowledge adds
nothing to static multi-agent epistemic logic, since its reduction axioms
eliminate every announcement operator. What it adds is a natural and succinct
way of \emph{stating} updates, and in the classroom that is what counts.} On
top of multi-agent epistemic logic (one knowledge operator per agent, whose
accessible worlds are those the agent cannot tell apart from the actual one),
PAL adds an operator $[!\varphi]\psi$, read ``after $\varphi$ is truthfully
announced, $\psi$ holds''. Its meaning is simple: after the announcement, only
the worlds in which $\varphi$ was true remain in play. In LogiKEy this is once
more a shallow embedding: knowledge is the box of Sect.~\ref{subsec:sse}, and
an announcement restricts all subsequent evaluation to the surviving worlds.
Technically this changes the type of a proposition once more: where a modal
proposition was a function from worlds to truth values,
$\sigma = i \Rightarrow \mathit{bool}$, a dynamic proposition takes one
further argument, the set of worlds currently in play,
$\tau = (i \Rightarrow \mathit{bool}) \Rightarrow i \Rightarrow \mathit{bool}$;
it is this extra argument that lets the same formula be evaluated before and
after an announcement. The treatment extends to common knowledge, and the
embedding has been proven sound in HOL~\cite{J58}
(Fig.~\ref{fig:pal}).

\begin{figure}[t]
\begin{lstlisting}[style=isa]
type_synonym tau = "(i => bool) => i => bool"   (* dynamic proposition: worlds in play, world *)
consts K   :: "agent => tau => tau" ("K_ _")     (* K_a p : agent a knows p  (S5) *)
consts C   :: "tau => tau"                       (* common knowledge among a, b, c *)
consts ann :: "tau => tau => tau" ("[!_]_")      (* [!phi]psi : after announcing phi, psi *)
consts ws  :: "agent => tau"                     (* ws x : x has a white spot *)
axiomatization where WM1: "$\val{C\ (ws\ a \oor ws\ b \oor ws\ c)}$"  (* at least one white *)
(* WM2: if x has no white spot, the other two know it (common knowledge; six axioms) *)
theorem "$\val{[!\neg K_a\ (ws\ a)]\ [!\neg K_b\ (ws\ b)]\ (K_c\ (ws\ c))}$"
\end{lstlisting}
\caption{Public-announcement logic in LogiKEy, in outline (the encoding
of~\cite{J58}, entered into the Archive of Formal Proofs). The operator \texttt{K}
is an $\mathbf{S5}$ box and stands for individual knowledge, while \texttt{C} stands for common knowledge in a group; the announcement
\texttt{[!$\varphi$]} restricts the model to the worlds where $\varphi$ held.
The king's statement is common knowledge (\texttt{WM1}), as is that each wise
man sees the others' spots (\texttt{WM2}). The puzzle is then the single
theorem shown: after $a$ announces that he does not know his colour, and $b$,
having heard him, announces the same, $c$ knows that he has a white spot.}
\label{fig:pal}
\end{figure}

With the embedding in place, the puzzle becomes a single theorem: given the
common knowledge that at least one hat is white, after the first wise man
announces that he does not know his colour and the second, having heard him,
announces the same, the third knows his spot is white. The automated provers
discharge it, and, just as
instructive for students, Nitpick can display the intermediate epistemic
states, showing how each announcement prunes possibilities until only one
survives. The same machinery settles variants (two wise men, black spots, a
false announcement) by proof or by countermodel.

This closes the loop opened in Rung~1. The single escalation
\emph{propositional $\to$ modal $\to$ dynamic epistemic} (each step motivated
by a demonstrated limitation of the previous logic in modelling ``says'' and
``knows'') is, in miniature, the whole argument of the chapter: many logics,
one meta-logic, each one \emph{earned} rather than presented.

\subsection*{Interlude---The virtues of \emph{higher-order} logic: Boolos's inference}
\label{subsec:boolos}

By now a sharp student will be asking what is gained by taking the meta-logic
to be \emph{higher-order} rather than ordinary first-order logic. One example
illustrates it memorably, and does so in a domain where one might least expect
expressiveness to help: automated proof search~\cite{J61}. Boolos's ``curious
inference''~\cite{Boolos1987} is a small, valid argument (five axioms in all:
three characterising a fast-growing, Ackermann-style function, plus a base case
and an induction-like step for a predicate) whose conclusion, that this
predicate holds of a particular term, genuinely \emph{follows}. Three claims
should be kept apart here. First, a theorem about \emph{proof length}: Boolos
showed that in a \emph{cut-free} first-order calculus the shortest derivation
has on the order of $f(5,5)$ steps, for $f$ the Ackermann function, so that no
such proof could be found, or even written down, in this
universe~\cite{Boolos1987}. The restriction to cut-free calculi is not an
idealisation but the practically relevant case, since the resolution,
superposition and tableau calculi on which automated provers are built are
analytic. Nor would admitting cuts rescue the first-order case: the cut formula
one would need is an instance of second-order comprehension, which quantifies
over properties and is therefore not expressible in first-order logic at all. Second, a theorem about \emph{expressiveness}:
Boolos also gave a short proof in second-order logic, using particular
instances of the comprehension schema~\cite{Boolos1987}. Third, an
\emph{observation} about today's provers: they are built around cut
elimination and do (generally) not introduce the comprehension instances that
would be needed. Higher-order provers are not
quite in the first-order predicament here, since $\lambda$-abstraction together
with primitive substitution addresses \emph{cut introduction} in part, but in general
this remains far too weak. Once a single higher-order definition is supplied by
hand (the notion of a property being ``inductive'', which quantifies
over \emph{predicates}), the provers synthesise the remaining lemmas and
dispatch the theorem in milliseconds~\cite{J61}. The moral cuts against the grain of what a
theoretical-computer-science course tends to instil. Complexity results teach
that greater expressiveness must be \emph{paid} for, with undecidability or
higher complexity; yet here the \emph{more} expressive logic affords an
incomparably \emph{shorter} proof. Expressive power is not merely a luxury;
for some problems it is what brings a proof within reach at
all.

This proof-theoretic bonus is, however, only the most dramatic reason for
choosing HOL as the base layer, not the primary one. The primary reason is that
higher-order logic is the natural \emph{home} for the semantical embeddings approach.
The $\lambda$-abstraction of~\eqref{eq:sse} lifts each connective to a
world-dependent function, and quantification over predicates lets one speak
directly about properties and accessibility relations. HOL's mature provers
and model finders then serve every embedded logic for free.

\subsection{Rung 3---Reasoning about norms: Chisholm's paradox}
\label{subsec:rung3}

With one modality in hand, we reinterpret it deontically ($\Box$ as ``it
ought to be that'') and arrive at so-called ``standard deontic logic'' (SDL), the modal
logic $\mathbf{KD}$. The vehicle is Chisholm's contrary-to-duty paradox~\cite{Chisholm1963}, a small set of intuitively consistent obligations that
classical SDL formalisations render inconsistent or redundant~\cite{HandbookDeontic1}.
Two things make this rung pedagogically decisive. First, it shows non-classical
logic doing genuine work on a problem philosophers care about. Second, it is
where \emph{countermodels become pictures}: a raw Nitpick countermodel, all
but unreadable as text, is rendered automatically as a labelled Kripke graph
by a lightweight visualiser developed within LogiKEy~\cite{W72} (Sect.~\ref{sec:tools}). The student does not
merely read ``not valid''; she gets a visual explanation of \emph{why} the
formula fails.

\begin{figure}[t]
\centering
\begin{tikzpicture}[>=Stealth,node distance=14mm,font=\scriptsize,
   world/.style={circle,draw,minimum size=6.5mm,inner sep=0pt},
   lbl/.style={font=\scriptsize,inner sep=1pt}]
  \node[world] (w1) {$w_1$};
  \node[world,right=of w1] (w0) {$w_0$};
  \node[world,right=of w0] (w2) {$w_2$};
  \node[lbl,above=5mm of w0] (s) {start};
  \draw[->] (s) -- (w0);
  \draw[->] (w1) -- (w0);
  \draw[->] (w0) -- (w2);
  \draw[->] (w2) to[out=-40,in=40,looseness=6] (w2);
  \node[lbl,below=0.5mm of w1] {$go,tell$};
  \node[lbl,below=0.5mm of w0] {$go,tell$};
  \node[lbl,below=0.5mm of w2] {$go,tell$};
\end{tikzpicture}
\caption{The Nitpick countermodel of the text, rendered automatically as a
labelled Kripke graph by the LogiKEy visualiser~\cite{W72}: every world
satisfies $g$ and $t$ (``$go,tell$''), the arrows are the deontic accessibility
relation, and \emph{start} marks the world of evaluation. Reading the picture
beats reading Nitpick's raw text---that is the point of the visualiser.}
\label{fig:kripke}
\end{figure}
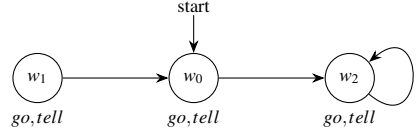

In ordinary language the four sentences are: (1)~it ought to be that Jones
goes to assist his neighbours; (2)~it ought to be that if he goes, he tells
them he is coming; (3)~if he does not go, he ought not to tell them he is
coming; (4)~he does not go. The four textbook SDL renderings of these
sentences differ only in whether each obligation takes wide (\textbf{W}) or
narrow (\textbf{N}) scope over the conditional; we take the four Isabelle/HOL encodings from the LogiKEy
Workbench~\cite{J51,W72}:
\begin{center}
\setlength{\tabcolsep}{11pt}\renewcommand{\arraystretch}{1.15}
\begin{tabular}{lllll}
\toprule
 & WN & WW & NN & NW\\
\midrule
F1 & $\mathbf{O}g$ & $\mathbf{O}g$ & $\mathbf{O}g$ & $\mathbf{O}g$\\
F2 & $\mathbf{O}(g \to t)$ & $\mathbf{O}(g \to t)$ & $g \to \mathbf{O}t$ & $g \to \mathbf{O}t$\\
F3 & $\neg g \to \mathbf{O}\neg t$ & $\mathbf{O}(\neg g \to \neg t)$ & $\neg g \to \mathbf{O}\neg t$ & $\mathbf{O}(\neg g \to \neg t)$\\
F4 & $\neg g$ & $\neg g$ & $\neg g$ & $\neg g$\\
\bottomrule
\end{tabular}
\end{center}
where $g$ reads ``Jones goes to help'' and $t$ ``Jones tells them he is
coming''. None of these renderings captures the intended meaning (each either
makes the four sentences jointly inconsistent or makes one a consequence of the
others), and students probe them by \emph{experiment}. One such check asks
whether the fact $F_4$ (that Jones does not go) already follows from the three
obligations $F_1$--$F_3$: it does not, and Nitpick returns a model in which
$F_1$, $F_2$, $F_3$ all hold yet Jones in fact \emph{goes}. That model, rendered
by the LogiKEy visualiser~\cite{W72}, is Fig.~\ref{fig:kripke}; because $g$ and $t$ hold at every world, each node
carries the label ``$go,tell$'', and the world marked \emph{start} is where the
conjectured entailment $F_1\wedge F_2\wedge F_3\to F_4$ fails.

The machine's verdicts diagnose the four renderings precisely: at the
designated world, WN is inconsistent; WW makes $F_3$ follow from $F_1$; NN
makes $F_2$ follow from $F_4$; and NW exhibits both
redundancies. This
systematic failure of SDL on Chisholm's scenario is, moreover, precisely
what motivates the move to more expressive deontic logics. \emph{Dyadic}
deontic logics use a conditional obligation $\mathbf{O}(t\mid g)$, ``$t$ is
obligatory given $g$''. In {\AA}qvist's systems~\cite{Aqvist2002} it is
interpreted over a \emph{preference} (betterness) relation on worlds rather
than a mere accessibility relation; in the system of Carmo and
Jones~\cite{CarmoJones2002}, over a semantics designed expressly for
contrary-to-duty scenarios. In LogiKEy this upgrade is one further embedding, not
a new tool: the LogiKEy Workbench provides ready encodings of both
traditions~\cite{J43,B21,J51}, and the visualiser of Sect.~\ref{sec:tools}
displays preference relations just as it does accessibility
relations~\cite{W72}. The failed formalisations thus do double duty in class:
they diagnose SDL, and they justify its successors.

Chisholm's puzzle is a doorway, not a destination. The same deontic machinery
scales to substantial applications in AI and law, and the course points to
them. LogiKEy has been used to model \emph{value-oriented} legal reasoning,
where competing legal values are weighed to justify a decision~\cite{J65}.
Closer to this chapter's authors, it has served to formalise \emph{epistemic
rights}, such as the ``power-right to know''~\cite{C101}, 
or \emph{mental privacy}~\cite{C107,C108},
whose formalisation combines the logics used for the right to privacy and freedom of thought.
The very contrary-to-duty pattern of Chisholm's
puzzle even recurs in real regulation: one such structure in the GDPR is singled
out, in the LogiKEy literature, as instructive ``also from an educational
viewpoint''~\cite{J48}. In the classroom these connect a
small paradox to live questions in machine ethics and AI \& law, and let a
project-minded student move from a textbook example to a fragment of a genuine
normative theory.

\subsection{Rung 4---G\"odel's ontological argument}
\label{subsec:rung4}

The capstone lifts everything to \emph{higher-order} modal logic and applies
it to a famous philosophical argument with roots in Leibniz and Anselm:
G\"odel's modal ontological proof. The
encoding is remarkably compact: a single primitive, ``positive property'',
and three definitions (God-like, essence, necessary existence) on top of the
higher-order modal embedding, followed by five axioms
(Fig.~\ref{fig:godel}).

\begin{figure}[t]
\begin{lstlisting}[style=isa]
consts posProp :: "(e => sigma) => sigma"  ("P")   (* 'P A': A is positive *)
definition God :: "e => sigma" ("G")   (* God-like: has every positive property *)
  where "G x = $(\ofa A.\ P\ A \oimp A\ x)$"
(* also Ess (essence), NE (necessary existence), and Scott's axioms A1..A5 *)
\end{lstlisting}
\caption{The core of Scott's variant of G\"odel's argument: one primitive
(the positive-property operator \texttt{P}) and the definition of God-likeness
(the bold $\ofa,\oimp$ are the \emph{embedded} object connectives). From this
base---three definitions and five axioms---the provers derive
$\val{\obox\,\oex x.\ \mathrm{God}\,x}$, ``a God-like being necessarily exists'', while
Nitpick confirms the axioms are consistent by exhibiting a model.}
\label{fig:godel}
\end{figure}

Encoded this way, the argument becomes an object of experiment. Automated
provers reconstruct Scott's chain of results (positive properties are
possibly instantiated, hence a God-like being is possible; being God-like is
an essential property; and hence a God-like being necessarily exists), while
Nitpick confirms at each stage that the axioms remain consistent by exhibiting
a model~\cite{J70}. Turned instead on G\"odel's
\emph{original} axioms, the very same tools expose a hidden inconsistency,
from which everything would follow: a discovery first made by the first
author's higher-order prover LEO-II~\cite{J29}, and repaired in Scott's
variant~\cite{C55}. The machine refutation even admits a human-readable
reconstruction: an ``empty essence'' lemma from which a contradiction follows
in a few steps, an issue that had gone unnoticed before~\cite{C55}. The encoding
likewise makes the modal \emph{frame conditions} explicit per result: the final
theorem needs a \emph{symmetric} accessibility relation, whereas the
inconsistency already bites in the base logic $\mathbf{K}$~\cite{J70}. We show
only the headers and report
the machine's verdicts; the point is not to grind through a proof but to
witness the same platform that ran the hour-one puzzle now adjudicating a
twentieth-century metaphysical argument.

For students this capstone does real pedagogical work beyond spectacle. It
shows that the tools they first met on a logic puzzle are the very ones now used
in research; the automated analysis of G\"odel's argument was itself a noted
episode of ``AI in metaphysics''~\cite{C55}. And it makes vivid that formalisation can \emph{clarify} a famous argument, exposing
assumptions and errors that centuries of informal scrutiny had passed over. That
a mixed group of undergraduates can, by the end of a term, replay and
interrogate such a result is, in our experience, the course's most persuasive
advertisement for logic itself.

\medskip\noindent
Taken as a whole, the ladder is itself the argument for pluralism-in-teaching:
one environment, a family of logics of increasing expressiveness, continuous
machine checking, and students from different disciplines meeting on one
platform.

The ladder is, moreover, not a fixed track, and student projects can reach
genuine research material, because the same platform carries the teaching
examples and the research encodings. A few examples from this line of courses
show how far that can go. A project on public-announcement logic grew, by way
of a thesis, into the embedding with common knowledge used in
Sect.~\ref{subsec:wisemen}, published in the \emph{Journal of Logic and
Computation}~\cite{J58} and entered into the Archive of Formal
Proofs~\cite{J55}. A project on epistemic rights became the dynamic logic of
the ``right to know'' mentioned in Rung~3~\cite{C101} and an analysis of the
logical modalities in the European AI Act~\cite{C97}. Doctoral work at the
University of Luxembourg produced the embeddings of dyadic deontic
logics~\cite{J43,B21} that form much of the LogiKEy Workbench~\cite{J51} on
which Rung~3 draws. A mathematics-leaning branch is open as well: one project
embedded positive \emph{free} higher-order logic in HOL~\cite{C83}, and another
built on such a free logic to develop category theory in Isabelle/HOL with Dana
Scott~\cite{J47,C92}. The rungs shown here are one instructive path through a
much larger space of examples.

\section{Tool Support That Lowers the Barrier}
\label{sec:tools}

Several of the tools have already appeared in passing; together they are
what makes the approach usable by students who are not, and need not become,
specialists in interactive theorem proving.

\runinhead{Proof and refutation companions.}
Sledgehammer~\cite{BlanchetteEtAl2016} and Nitpick~\cite{BlanchetteNipkow2010}
address the beginner's ``blank page'' problem from both sides: one looks for a
proof, the other for a counterexample (Isabelle's \texttt{try} command even
invokes both at once). A student who does not yet know whether
a statement is true can simply ask and, whatever the answer, receive
something to inspect and learn from.

\runinhead{From countermodel text to a picture.}
Nitpick reports its (counter\mbox{-})models as textual assignments
(Fig.~\ref{fig:nitpickraw}). These are readable enough once one is used to
them, but they become inconvenient when large, and they use a function-update
notation that newcomers rarely know. The
LogiKEy Kripke-model visualiser~\cite{W72} closes this gap:
it parses a Nitpick model and emits a labelled graph of worlds, valuations,
and accessibility or preference edges, ready for slides or papers. It lets
students of law and philosophy \emph{see} where obligations clash, and it is
freely available.\footnote{See
\url{https://logikey.org} and the tool sources referenced in~\cite{W72}.}
Related infrastructure is emerging in the TPTP world, independent of any one
prover: a standard format for interpretations that covers Kripke
models~\cite{SutcliffeEtAl2024TPTPInterpretations}, tools that verify and
visualise such interpretations~\cite{SteenEtAl2023Interpretations}, and finite
model finding for first-order modal logics~\cite{SariyantoSteenSutcliffe2026}.

\begin{figure}[t]
\begin{lstlisting}[style=isa]
Nitpick found a counterexample for card i = 3:
  Skolem constant:  w = i1                       (* the world marked 'start' *)
  Constants:
    go   = (%x. _)(i1 := True,  i2 := True,  i3 := True)
    tell = (%x. _)(i1 := True,  i2 := True,  i3 := True)
    (R)  = (%x. _)((i1,i3):=True, (i2,i1):=True, (i3,i3):=True, ...)
\end{lstlisting}
\caption{The raw Nitpick text (abridged) of the \emph{very} model that
Fig.~\ref{fig:kripke} displays: Nitpick's worlds \texttt{i1},\texttt{i2},%
\texttt{i3} become $w_0,w_1,w_2$, the \texttt{Skolem constant} \texttt{w=i1} is
the world marked \emph{start}, the \texttt{(R)} pairs become the arrows, and
\texttt{go}/\texttt{tell} (i.e.\ $g$ and $t$) hold at every world. What the
visualiser adds is legibility: the same content, but readable by a student of
law or philosophy.}
\label{fig:nitpickraw}
\end{figure}

\runinhead{Automated assessment.}
Because an Isabelle proof either checks or does not, submissions can be graded
by machine. In larger deployments this has been realised by coupling Isabelle
with the DOMjudge online judge, software originally written for programming
contests~\cite{C58}. A student uploads a proof document, it is checked against
the lecturer's specification, and a verdict is returned within minutes;
unsuccessful attempts are retained and annotated inside the verified document. This matters most where hand-marking runs out. In a large
course a student might otherwise receive only a single, final comment, whereas
the machine's verdict arrives \emph{before} submission and can steer the work
rather than merely grade it. Even in the small cohorts of our own courses, where
grading is done by hand, the same automatic checking sustains a rapid
conjecture--attempt--feedback loop during class.

\runinhead{Autoformalisation with language models.}
A more recent aid addresses the hardest step for most students, the passage
from an informal statement to a formal one. Large language models now produce
serviceable first drafts of a formalisation from a natural-language
description, and we already include a small in-class demonstration. What makes
this usable rather than hazardous is the setting: a draft is not accepted on
trust but handed to the proof assistant, and Sledgehammer and Nitpick report
whether it proves what was intended or admits an unintended countermodel.
Students must still judge whether the definitions and assumptions capture the
intended meaning. The formalise--test--revise loop of Sect.~\ref{sec:why} thus absorbs the language
model as one more, fallible, participant, and connects the course to current
questions about neuro-symbolic AI. The same technology also lets students
outsource assignments, a difficulty shared with computer science and
mathematics at large. In our setting, this shifts assessment towards students'
own understanding, assessed through discussion: explaining why a formalisation
admits a given countermodel, comparing two formalisations of the same text, and
defending a modelling choice.
At Bamberg this connection is now taught
deliberately: the undergraduate logic course \emph{Logical Knowledge
Representation and Reasoning} (AISE-LKR-B) is paired with a course on DevOps
for AI systems (AISE-DO-B) that introduces agentic infrastructure and builds
bridges between theorem provers, wrapped as agents, and generative AI tools.
Together the two courses have become a starting point for ambitious student
projects and thesis topics.

\runinhead{One honest caveat.}
Using a general-purpose \emph{research} proof assistant, rather than a
purpose-built teaching applet, buys power and continuity with research
practice (students work with the very tools used in current research), at the
price of a steeper initial interface. Section~\ref{sec:experience} weighs this
trade-off against the alternatives.

\section{Experience, and Related Approaches}
\label{sec:experience}

\runinhead{What we observed.}
Of the three kinds of claim distinguished in Sect.~\ref{sec:why}, the
classroom observations are the softest. Over the three Bamberg runs a few
things stood out.
\emph{Exploration engaged students across disciplines.} Being able to ask the
machine (does this hold, and what breaks it?) and to receive an immediate proof
or an explicit countermodel turned otherwise abstract questions into something
students could probe. The mixed groups often taught one another: the
computer-science students were typically quicker with the tool, the philosophy
students quicker to see what was at stake in an example.
\emph{The capstone motivated the climb.} Announcing at the outset that the
course would end by having the computer scrutinise G\"odel's argument gave the
earlier, drier material a visible purpose.
\emph{The main friction was the tool itself.} Isabelle's syntax, the gap
between \emph{finding} a proof with Sledgehammer and \emph{understanding} it,
and installation across heterogeneous laptops were recurring hurdles,
especially for students without a programming background; starting from very
small, editable examples was the most effective remedy.
\emph{Automation reshaped what we could ask.} Because the system checks every
step, assignments could be genuinely open-ended (``find a model in
which\ldots'', ``decide whether\ldots and justify your answer formally'')
rather than restricted to what is convenient to mark by hand.
\emph{Students became reviewers.} We have discussed research-level,
work-in-progress Isabelle theories in class, and more than once a student's
difficulty or question uncovered a genuine issue, leading to a corrected
definition or a missing premise being added.

\runinhead{Relation to other tool-based approaches.}
Tool support for teaching logic is by now a lively area with a survey of its
own~\cite{TranMinhEtAl2025}, and this volume itself gathers several
complementary tool-based proposals; we leave their comparison to the editors'
introduction and confine ourselves here to a selection of work published
elsewhere, with no claim to completeness.
Interactive proof assistants are increasingly used in the teaching of
mathematics and of program verification: Coq in the \emph{Software
Foundations} series~\cite{PierceSF} and, for many years, in Smolka's
\emph{Introduction to Computational Logic} at Saarland
University~\cite{SmolkaMPCTT}, Lean in Avigad, Lewis and van Doorn's
\emph{Logic and Proof}~\cite{AvigadLogicProof}, and Isabelle in dedicated
logic and automated-reasoning courses~\cite{VilladsenJacobsen2021}. A range of
browser-based tutors, such as Carnap~\cite{LeachKrouse2017} and
ProofWeb~\cite{Hendriks2010}, supports the learning of natural deduction
through a gentle, purpose-built interface. Philosophy departments in
particular have a long tradition of such software for classical logic, from
Barwise and Etchemendy's \emph{Language, Proof and Logic} with its Tarski's
World and Fitch programs~\cite{BarwiseEtchemendy1999} and Allen and Hand's
\emph{Logic Primer} with the Logic Daemon~\cite{AllenHand2001} to the Jape
proof calculator~\cite{BornatSufrin1999}; at the largest scale, Sieg's
\emph{Logic \& Proofs} course at Carnegie Mellon, built on the AProS
proof-search engine and its Proof Lab, has taught natural deduction to many
thousands of students over two decades~\cite{Sieg2007AProS,OLILogicProofs}.
Closer to the boundary between informal and formal that our language-model
aid addresses (Sect.~\ref{sec:tools}), the Naproche system developed at Bonn
checks proofs written in a controlled natural language, and its didactic
variant Diproche has been used in introduction-to-proof
classes~\cite{DeLonEtAl2021Naproche,CarlEtAl2022Diproche}. ProofBuddy takes a middle road,
placing a simplified web front end with restricted syntax in front of Isabelle
itself~\cite{KarstenEtAl2023ProofBuddy,KarstenEtAl2025ProofBuddy}. A further
strand gamifies the assistant outright: Buzzard's \emph{Natural Number Game}
walks students into Lean through a sequence of levels, and a dedicated game
server now hosts a growing family of such games~\cite{NaturalNumberGame}, a
format that ties this chapter to the games discussed elsewhere in the present
volume. Ontology
tools in the style of Prot\'eg\'e introduce logic through applications; even
that terrain is not foreign to our course, since the basic description logic
ALC is, by Schild's classic observation~\cite{Schild1991},
propositional multimodal logic in another notation. In their \emph{semantical embedding} this
correspondence becomes a one-line exercise, and the soundness of the usual
ALC tableau rules a small verification task, as we showed in an
earlier tutorial~\cite{C46}; its concepts thus fall out of the modal
embedding of Rung~2, a connection we make explicit in the lecture when time
permits. LogiKEy's place in this
landscape is characterised by two choices. First,
where most such tools fix a single logic or calculus, LogiKEy is
\emph{semantics}-based and \emph{logic-pluralistic}: many, mostly
non-classical, object logics are obtained by embedding their semantics, so
that each of them comes with both proof search and countermodel search. Second, it deliberately repurposes a general-purpose
\emph{research} proof assistant, accepting a steeper learning curve in
exchange for greater reach: the same environment carries a
first logic puzzle and a research-grade
metaphysical argument, and it does so for a mixed audience of computer
science, mathematics, and philosophy students. The trade-off is better described as an
exchange than as a loss: the aim is not to eliminate formal overhead, but to
exchange the repeated, ad-hoc formalisation costs of teaching one logic at a
time for a single reusable infrastructure. A purpose-built tutor is an excellent first encounter with formal
proof, and a LogiKEy-style course is a natural follow-up for students who want
to go further.

\section{Conclusion and Outlook}
\label{sec:outlook}

We have argued that a single expressive meta-logic, hosted in a modern proof
assistant, lets a mixed student cohort learn many logics uniformly and
experimentally: one meta-logic, many object logics, always machine-checked.
We have illustrated the claim with a graded ladder of classroom examples and
with the tools that make it workable.

It is worth being precise about what the chapter establishes and what it
proposes. \emph{Established}, in the sense of machine-checked and reproducible
from the openly available sources, are the formal results of the ladder and
the methodological claim that one meta-logic carries all of them.
\emph{Documented}, but only by our own experience, is that a course built this
way can be run with a mixed cohort and that students engage with it in the ways
described in Sect.~\ref{sec:experience}. \emph{Open} are the questions a
pedagogical study would have to settle: whether the approach improves
particular skills, whether it lowers or raises barriers for particular groups,
how it compares with purpose-built tutors on a common task, and how learning is
best assessed now that large language models can draft formalisations. We regard the
chapter as an invitation to such studies, not as a substitute for them.

Several directions look promising. Language-model assistance with
formalisation (Sect.~\ref{sec:tools}) suggests a future in which students
move still more fluently between informal argument and machine-checked proof.
Richer interactive and visual tooling, in particular
interactive exploration of Kripke models extending the visualiser
of~\cite{W72}, promises to make non-classical semantics more
tangible yet.

A further direction concerns \emph{portability across proof assistants}.
Because LogiKEy is defined at the level of a \emph{meta-logic} rather than of a
particular tool~\cite{J40,C90}, the approach
is not wedded to Isabelle: any proof assistant whose logic extends higher-order
logic can host the same semantical embeddings. We have explored this by porting
the ontological-argument material of Rung~4 to the Lean~4 proof
assistant~\cite{BenzmuellerLean2026}. The translation is structure-preserving,
reproducing the same embeddings, definitions, and theorems (down to the
inconsistency of G\"odel's original axioms and the individual frame conditions
on which each result depends), and the complete Lean sources are available with
the preprint. Such portability matters
pedagogically: it lets a course meet students on whichever proof assistant
their programme already uses.

The ceiling, moreover, is high. Recent work carries the embedding technique,
with \emph{machine-checked faithfulness}, to monadic second-order
logic~\cite{B31} and even to a deep embedding of HOL within HOL whose
soundness, completeness, and consistency are themselves verified~\cite{J74afp},
so that a graduate course can treat \emph{metatheory itself} formally. A
growing library of openly available, verified developments offers instructors
ready-made, machine-checked material at whatever level they
teach.\footnote{See the authors' entries in the Archive of Formal Proofs,
\url{https://isa-afp.org/authors/benzmueller/}.}

Finally, the same platform that carries our teaching also carries current
research on normative and epistemic reasoning, which points towards a natural
extension of the approach to the teaching of AI ethics. All of the course
materials are openly available,\footnote{Via \url{https://logikey.org} and the
associated Isabelle/HOL datasets.} in the hope that others may adapt them, in
keeping with the mission of the Commission on Logic Education from which this
volume grows.

\begin{acknowledgement}
We are grateful to the LogiKEy contributors mentioned in this chapter and
listed at \url{https://logikey.org}, and to the students of the courses
reported here, whose questions shaped much of the material.
The work of Luca Pasetto was supported by the Luxembourg National Research Fund (FNR)
through the project Logical methods for Deontic Explanations (INTER/DFG/23/17415164/LODEX).
\end{acknowledgement}

\printbibliography

\end{document}